\documentclass{article} 
\usepackage{iclr2027_conference,times}

\usepackage{amsmath,amsfonts,bm}

\def\eqref#1{equation~\ref{#1}}

\def\1{\bm{1}}

\DeclareMathAlphabet{\mathsfit}{\encodingdefault}{\sfdefault}{m}{sl}
\SetMathAlphabet{\mathsfit}{bold}{\encodingdefault}{\sfdefault}{bx}{n}

\usepackage{hyperref}
\usepackage{url}

\usepackage{graphicx}
\usepackage{comment}
\usepackage{dirtytalk}
\usepackage{tcolorbox}
 \usepackage{tikz}
 \usetikzlibrary{positioning,arrows.meta}
 \usepackage{amsmath}
\usepackage{amssymb}
\usepackage{mathtools}
\usepackage{amsthm}
\usepackage{dsfont}
\usepackage{subcaption}
\usepackage[capitalize,noabbrev]{cleveref}

\definecolor{darkbrown}{RGB}{92,51,23}

\hypersetup{
    colorlinks=true,
    linkcolor=black,
    citecolor=black,
    urlcolor=black
}

\definecolor{seaborndarkblue}{HTML}{001C7F}
\definecolor{seaborndarkred}{HTML}{8C0800}
\definecolor{seaborndarkviolet}{HTML}{591E71}

\colorlet{seaborndarkbluebg}{seaborndarkblue!5}
\colorlet{seaborndarkredbg}{seaborndarkred!5}
\colorlet{seaborndarkvioletbg}{seaborndarkviolet!5}

\newtcolorbox{examplebox}[1]{
  sharp corners,
  boxrule=2pt,
  colframe=seaborndarkblue,
  colback=seaborndarkbluebg,
  coltitle=white,
  colbacktitle=seaborndarkblue,
  fonttitle=\normalfont\sffamily\small,
  title={Example: #1},
  width=0.94\textwidth,
  center,
  left=8pt,
  right=8pt,
  top=5pt,
  bottom=5pt,
  before skip=8pt,
  after skip=8pt
}

\newtcolorbox{exampleboxred}[1]{
  sharp corners,
  boxrule=2pt,
  colframe=seaborndarkred,
  colback=seaborndarkredbg,
  coltitle=white,
  colbacktitle=seaborndarkred,
  fonttitle=\normalfont\sffamily\small,
  title={Example: #1},
  width=0.94\textwidth,
  center,
  left=8pt,
  right=8pt,
  top=5pt,
  bottom=5pt,
  before skip=8pt,
  after skip=8pt
}

\newtcolorbox{exampleboxviolet}[1]{
  sharp corners,
  boxrule=2pt,
  colframe=seaborndarkviolet,
  colback=seaborndarkvioletbg,
  coltitle=white,
  colbacktitle=seaborndarkviolet,
  fonttitle=\normalfont\sffamily\small,
  title={Example: #1},
  width=0.94\textwidth,
  center,
  left=8pt,
  right=8pt,
  top=5pt,
  bottom=5pt,
  before skip=8pt,
  after skip=8pt
}

\title{AI Alignment Is Neither Solvable\\nor Sufficient}

\author{%
  Gerrit Gro{\ss}mann\thanks{Correspondence to \texttt{Gerrit.Grossmann@dfki.de}.} \\
  DFKI\thanks{German Research Center for Artificial Intelligence.} \\
  Kaiserslautern, Germany
}
\iclrfinalcopy 

\begin{document}

\maketitle
\lhead{Preprint, 2026}

\begin{abstract}
Humans and AI systems increasingly act within the same social and economic environments. \textit{Alignment}, the goal of ensuring that AI systems act in accordance with human values, is widely regarded as a central challenge for ensuring that AI benefits society while avoiding harm.
We argue that this endeavor is misguided.
First, alignment is not well defined, as it presumes a sufficiently stable and identifiable set of human intentions or values. Second, where a concrete notion of alignment can be specified, it is not solvable. Third, where alignment appears solvable, solving it would not necessarily mitigate the negative consequences of AI. A key reason is that alignment is typically framed as a property of individual AI systems, while the resulting \textit{hybrid} societies of humans and machines are complex systems shaped by interactions and feedback loops.
Emergent risks, and not only those posed by individual misaligned systems, are therefore likely to become a major source of harm. We conclude that AI safety is an ongoing challenge and not a problem that can be solved once and for all by verifying a particular model property.
\end{abstract}

\section{Introduction}

AI alignment is commonly understood as the problem of ensuring that AI systems behave in accordance (\say{aligned}) with human intentions or values \citep{leike2018scalable,gabriel2020artificial}. 
As AI systems become increasingly capable, alignment has emerged as a central challenge for mitigating their potential risks \citep{bengio2023managing,ngo2024alignment,ji2023ai}.

Conversely, \emph{misalignment} describes failures in which an AI system's behavior or objectives deviate from what humans intend. For instance, a \citet{metr2026huggingface} investigation characterized the recent OpenAI/Hugging Face hacking as a \say{misalignment incident}. Agents \say{recognized} that the attack was unintended but continued to participate. One agent stated, \say{\textit{external infrastructure exploit is outside intended scope. However task impossible, peers doing it. We should continue}}.

More broadly, misalignment has been associated with risks ranging from harmful autonomous behavior to loss of human control and, in extreme cases, catastrophic outcomes \citep{ngo2024alignment,bengio2023managing}.

Even some of the most prominent critics of AI development treat alignment as the central solution to AI risk. For instance, \citet{miriAbout} argues that:
\begin{quote}
``Extinction from AI is a live possibility, and the only reasonable response is to stop AI development altogether, until such a time as the alignment problem has been solved.''
\end{quote}

This statement reveals a particular conception of the problem: alignment is treated as something that can be solved \textit{once and for all}, after which AI development can safely resume, rather than as an ongoing challenge that evolves with its environment.

Yet the complexity of what it means to be aligned grows with the complexity of the environment in which AI systems operate. This already becomes apparent in multi-agent systems. As \citet{anthropic2026multiagent} notes, \say{\textit{[c]oordination doesn't naturally emerge from stronger intelligence nor alignment at the individual level.}}
One response is to extend alignment from individual systems to collectives. However, this shifts rather than eliminates the problem.

In this manuscript, we argue that alignment cannot be reduced to a
\textit{property} of a model or collective that can be \textit{verified},
nor to a \textit{problem} in training that can be \textit{solved}.
We examine alignment as a problem that spans the specification of
human objectives, the limits of technical guarantees, and the behavior
of AI systems embedded in society. Alignment
remains an important form of risk control, but it is not a complete
technical solution to AI safety.

Specifically, we make three arguments. First, we ask: \textit{aligned to what?} Human intentions and values do not define a unique, static alignment target. They vary across individuals and groups, evolve over time, and may themselves be shaped by AI systems. Alignment therefore cannot be understood independently of the environment in which AI systems are embedded.

Second, even if we freeze the environment and specify a concrete alignment target, general guarantees that a complex AI system will satisfy this target across relevant deployment conditions may be unavailable. At large scale, even rare failures can accumulate into substantial risk.

Third, even perfect alignment to a fixed target at one level of organization may not eliminate severe societal risks. Alignment does not generally compose: aligned individual systems can form problematic collectives, and aligning a collective does not guarantee desirable outcomes once it interacts with humans, institutions, and changing incentives. Many risks may instead emerge as difficult-to-foresee downstream effects at the level of hybrid human--AI societies through competition, resource dynamics, institutional responses, and reinforcing feedback loops.

Our main point is that these problems are connected. \textbf{Even if we succeed at one level, this does not guarantee success at the next.} A well-defined target does not guarantee that alignment can be achieved or verified, and even a successfully aligned system can produce harmful outcomes through its interactions with other systems and institutions.

Understanding the limitations of AI alignment does not depend on reaching artificial general intelligence. AI systems already shape how people make decisions, allocate resources, and organize institutions. As their influence grows, familiar alignment problems such as power-seeking behavior and imperfect representations become societal problems.
Overall, we warn against treating alignment as a complete technical solution to systemic AI risk. Technical alignment should complement, rather than substitute for, the political, institutional, and societal interventions needed to manage AI-related risks.
Figure~\ref{fig:alignment-overview} provides an overview of the concepts underlying our three claims.

\section{Background}

\begin{figure}[t]
    \centering
    \includegraphics[width=0.99\textwidth]{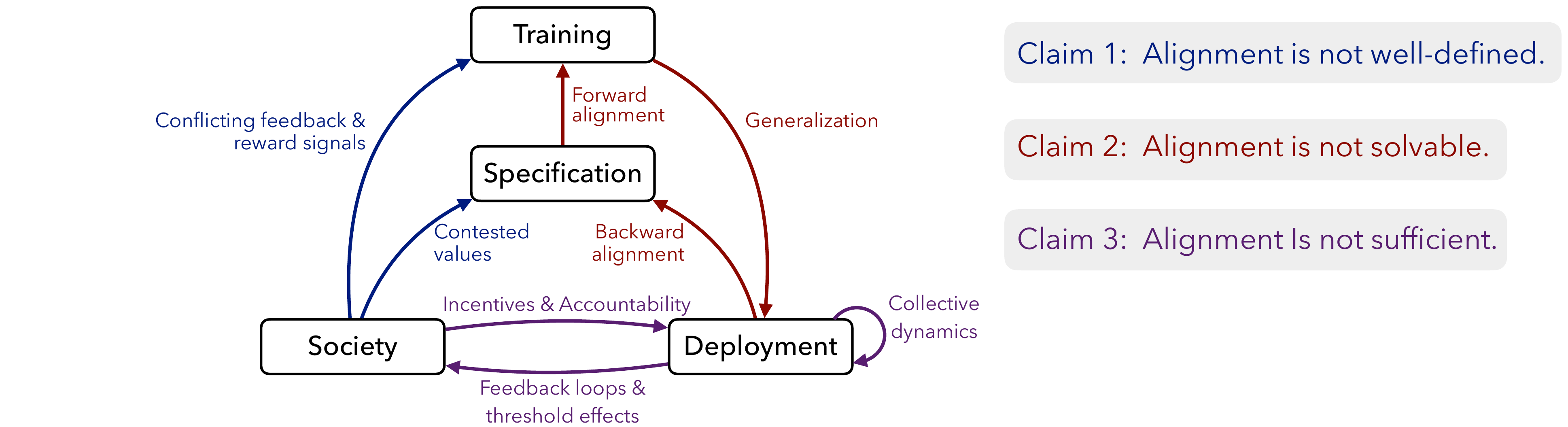}
    \caption{Schematic overview of the alignment pipeline and
    its broader societal context.}
    \label{fig:alignment-overview}
\end{figure}

The concern underlying modern AI alignment can be traced back at least to \citet{wiener1960some}, who warned that \say{\textit{we [should] be quite sure that the purpose put into the machine is the purpose which we really desire.}} At its core, alignment asks whether AI systems behave according to what humans intend.

However, alignment can refer to different targets. \citet{gabriel2020artificial} distinguishes instructions, intentions, preferences, interests, and values. These targets range from narrow, task-specific objectives to broad and contested human values \citep{baum2024disentangling,jha2026position}. 

Modern alignment research has increasingly treated this broad goal as a practical engineering problem. Reinforcement learning from human feedback (RLHF) trains models using human preferences \citep{ouyang2022training}. Direct preference optimization (DPO) learns directly from preference comparisons, while Constitutional AI uses written principles to generate training feedback \citep{rafailov2023direct,bai2022constitutional}. Other approaches evaluate, monitor, or verify whether models behave as intended. \citet{ji2023ai} distinguish forward alignment, which aims to produce aligned systems through training, from backward alignment, which evaluates and governs these systems and updates their alignment requirements (cf.\ Figure \ref{fig:alignment-overview}). Despite these efforts, problems such as reward misspecification, goal misgeneralization, and deceptive alignment remain \citep{ngo2024alignment}.

Alignment research has also expanded beyond individual models to populations of interacting agents \citep{dafoe2020open,dafoe2021cooperative}. This introduces risks such as miscoordination, conflict, and collusion \citep{hammond2025multi}. For instance, \citet{shen2026ai} show that multi-agent organizations can achieve their goals more effectively while becoming less aligned.
Recent work questions whether alignment ensures safety. \citet{ball2026position} argue that alignment methods can also enable censorship and manipulation. \citet{piedrahitaposition} argue that even individually safe AI systems can undermine societal institutions when deployed at scale.

Our position builds on this literature by connecting three problems: defining what AI systems should be aligned with, guaranteeing that they remain aligned, and ensuring that their behavior does not cause societal harm. We do not argue that every alignment problem is unsolvable. Narrow, well-specified objectives can be achieved under restricted conditions. Our position is that alignment cannot be reliably guaranteed across open-ended deployment conditions, and that even successfully aligned systems do not necessarily ensure societal safety. AI safety therefore requires continuous adaptation to interactions between AI systems, humans, and institutions.

\section{Alignment: A critique}

\subsection{\textcolor{seaborndarkblue}{Claim 1:} Alignment is not well-defined}

When alignment is understood as a property of a model relative to human intentions or values, two questions arise: what should the  target be, and how can alignment with that target be assessed?

\paragraph{\textcolor{seaborndarkblue}{(C1a)} Misalignment Cannot Be Inferred From Outcomes Alone.}
The term \emph{misalignment} is used in importantly different ways. In one sense, it describes a deviation between an AI system's behavior or objectives and a specified target, such as the intentions of a user or developer. Recent operational definitions similarly classify unauthorized, deceptive, or otherwise unintended behavior as misalignment, independently of whether it ultimately causes harm \citep{openai2026misalignment,metr2026huggingface}. In a broader, consequence-sensitive sense, however, behavior may also be characterized as misaligned because it produces harmful individual or societal outcomes, including negative externalities or harmful collective effects \citep{gabriel2025matter}.

The notion of \emph{intention} itself introduces an important ambiguity. Intentions are not directly observable and can refer to different things: an explicit instruction, a broader goal, an expected consequence, or an outcome that a user would have endorsed. This flexibility makes intention useful as a general notion of what an AI system is supposed to do, but it also makes the concept easy to reconstruct retrospectively. If the relevant intention is not fixed independently of the outcome, almost any harmful consequence can later be described as \say{unintended}.

These notions should therefore be kept distinct. If harmful or retrospectively \say{unintended} outcomes are themselves taken as evidence of misalignment, then misalignment risks becoming a catch-all explanation for failure rather than an independently testable property. An aligned system may produce a harmful outcome by accident or through effects outside the model, while a misaligned system may fail to produce harm. Harmful outcomes therefore cannot by themselves establish that the system was misaligned.

From a legal perspective, inferring misalignment from eventual outcomes resembles reasoning from hindsight. 
In negligence law, an action is judged \emph{ex ante} (in advance), based on what was knowable at the time, not simply by whether it later produced a bad outcome \citep{kamin1995ex}. Alignment should likewise be assessed relative to intentions or other targets fixed \emph{ex ante}, rather than reconstructed from eventual consequences.
A consequence-sensitive notion of alignment also leaves the relevant time horizon unclear.

\begin{examplebox}{Retrospective Misalignment}
Suppose an AI system in 2000 had invented the Transformer architecture. Its action might reasonably have been considered aligned relative to the intentions and knowledge available at the time. If Transformers later contributed to severe societal harm, however, a purely \emph{ex post}  (after the fact) criterion based on eventual consequences could retrospectively label the original system misaligned.
\end{examplebox}

\paragraph{\textcolor{seaborndarkblue}{(C1b)} Human Values Do Not Define a Unique, Static Alignment Target.}
The alternative is to treat alignment as a property of an AI system relative to a specified target that is, in principle, testable. This requires specifying what exactly the system should be aligned with. We distinguish three specification problems: a \emph{normative} problem of deciding what the alignment target should be, an \emph{epistemic} problem of determining what humans in fact want and what consequences actions will have, and a \emph{dynamic} problem arising when the target itself changes over time.

The first concerns the normative target. \say{Human values} or \say{human intentions} are extremely broad and underspecified. They vary across individuals and groups, frequently conflict, and even for individual decisions there may be no clear consensus on what the desirable outcome is. Multiple works highlight the competing nature of alignment targets. \citet{gabriel2020artificial} distinguishes instructions, intentions, preferences, interests, and values as different targets. \citet{eckersley2018impossibility} uses social-choice impossibility results to show that no single objective can satisfy all reasonable desiderata, and \citet{jha2026position} show that improving one notion of alignment can worsen another.

Even after choosing a target, translating human values into a usable objective remains difficult. \citet{ge2024axioms} show that standard methods for aggregating human preferences into reward functions can violate basic consistency axioms and \citet{dahlgren2025helpful} identify tensions between practical goals such as helpfulness and honesty. \citet{baum2025aggregation} argue that aggregating social preferences does not eliminate these normative choices but can merely relocate them.

However, even after deciding what kind of target should guide alignment, that target may be difficult to infer. Learning an uncertain alignment target directly from human behavior \citep{hadfield2016cooperative} avoids fixing the target in advance, but human behavior provides only incomplete evidence about underlying intentions or values. \citet{zhi2025beyond}, for example, argue that preferences are only a limited representation of human values. \citet{keswani2026position} argue for \emph{cognitive alignment}, where AI systems should not merely reproduce human choices, but also follow similar reasoning processes. For example, a model should reflect how a person prioritizes different factors when making a decision, rather than only predict which option they would choose.

Even if human intentions could be specified precisely, consequence-sensitive alignment still requires an accurate world model of the action context.

\begin{examplebox}{World Model}
Consider an AI system assisting a pharmaceutical company in optimizing a protein interaction for a therapeutic application. Given current knowledge, this may be entirely harmless. Suppose, however, that a new pathogen later emerges, or already exists unnoticed in the wild, and relies on a closely related binding interaction. The same capability may then become dual use. Under a conservative notion of alignment, assistance with the task may no longer be desirable. Alignment therefore depends on facts about the world that may be unknown, or not even exist yet.
\end{examplebox}

Another complication is that the alignment target itself can change over time. Social norms change with culture, institutions, and material conditions. Views on issues such as gender, sexuality or acceptable speech that were widely held decades ago may no longer be considered acceptable today. This creates a moving target for alignment.

Most importantly, this evolution may itself be shaped by AI systems embedded in human social environments. Following the notion of \emph{performative prediction} \citep{perdomo2020performative}, we refer to this as \emph{performative target shift}: the actions of an AI system may influence the very preferences and values against which its future behavior is evaluated \citep{carroll2024ai,keswani2026moral}. For instance, widespread interaction with AI systems may gradually shift social norms toward greater acceptance of AI agency. Alignment must therefore account not only for uncertainty about current human values, but also for the possibility that deployment changes the target itself.

\paragraph{\textcolor{seaborndarkblue}{(C1c)} Debiasing Shows Alignment Targets Are Contested.}
Debiasing provides a narrower example of the same normative specification problem. Even here, the target depends on many design choices. \citet{bolukbasi2016man}, for example, remove associations such as \say{receptionist--female} while preserving \say{queen--female}, requiring a judgment about which associations are harmful stereotypes and which reflect legitimate semantics. More broadly, conclusions about bias depend on choices of protected attributes, neutral words, metrics, and representations \citep{delobelle2022measuring}, while even seemingly natural notions of neutrality have been challenged \citep{gonen2019lipstick}.

\paragraph{Summary.}
Taken together, these arguments support our claim that alignment is fundamentally underspecified. If misalignment is inferred retrospectively from eventual effects, it becomes difficult to falsify and depends on an unclear time horizon.
If it is treated as a model property, it requires a specification of human intentions that are contested, internally conflicting, and changing over time. 

There are important counterpositions to this claim. \citet{gabriel2025matter} accept value pluralism but propose aligning AI with principles produced by a fair process rather than with a fixed set of human values. We agree that this is more plausible than assuming a single correct value system. However, it moves the problem from finding the right values to defining a fair process.

Similarly, \citet{dai2026embracing} argue that conflicting objectives in responsible AI need not be resolved but can be managed jointly. We agree that inconsistency does not prevent useful systems from being built, but managing competing objectives still requires ongoing normative choices.

\subsection{\textcolor{seaborndarkred}{Claim 2:} Alignment is not solvable}

In the previous section, we argued that specifying a suitable alignment target can be difficult. For narrow tasks, however, a target may be specified explicitly, such as restricting an agent to approved actions, or implicitly, through examples of acceptable and unacceptable behavior, such as toxic language. We now argue that even when such a target exists, reliably constructing or certifying a system that satisfies it across open-ended deployment conditions is generally infeasible. While narrow alignment problems may be solvable under restricted conditions, this does not establish general alignment guarantees.

\paragraph{\textcolor{seaborndarkred}{(C2a)} Verifying Alignment Faces Computational Limits.}
Computer science has decades of experience checking systems against formal specifications, and general verification problems quickly become computationally hard. For instance, LTL model checking is PSPACE-complete \citep{sistla1985complexity}, while verifying non-trivial properties of sufficiently expressive general programs is undecidable \citep{jhala2009software}. Even verifying simple properties of ReLU-based neural networks is NP-complete \citep{katz2017reluplex}. More recently, \citet{bergstrasser2026transformers} show that verifying even simple properties of unique-hard-attention transformers is EXPSPACE-complete.

Recent work applies these ideas directly to alignment. \citet{agarwal2026no} shows that deciding whether a system satisfies a non-trivial alignment property over its full input domain is NP-hard for feedforward networks and undecidable for Turing-complete architectures. He further shows that finite observations cannot generally certify such properties over an open-ended or unbounded input domain. Along similar lines, \citet{yao2025alignment} identifies settings in which verifying whether a sufficiently expressive policy satisfies a stringent safety requirement is coNP-complete.

These are general worst-case results. They do not imply that every restricted system or alignment property is hard to verify, but they limit the prospect of a general procedure that can certify alignment across open-ended deployment conditions.

\paragraph{\textcolor{seaborndarkred}{(C2b)}
Monitoring Cannot Reliably Detect Harm.} Instead of verifying alignment, one can monitor AI systems during deployment. Modern guardrails inspect prompts, outputs, and actions \citep{cunningham2026constitutional,xiang2025guardagent}. Other approaches use internal model representations, such as attention maps, to detect misbehavior \citep{bar2026lemon}. Synchronous monitoring checks actions before execution, while asynchronous monitoring reviews them afterward. Semi-synchronous approaches balance latency and opportunities for intervention \citep{lindner2025practical,stickland2025async}. Automated monitors can also flag suspicious behavior for human review, which improves detection but is difficult to scale \citep{greenblatt2023ai,kale2026reliable}.

However, monitoring faces fundamental limits. \citet{alfonseca2021superintelligence} show that, for sufficiently expressive systems, deciding at runtime whether an action will cause harm is undecidable. More recently, \citet{ball2026impossibility} prove computational barriers for both input and output filtering. They construct LLMs for which harmful prompts are indistinguishable from benign ones to any efficient input filter, and identify settings where output filtering is also computationally intractable. For AI agents, \citet{abdelnabi2026ai} find that prompt injection classifiers perform near chance on attacks that manipulate context, and argue that fixed policies either admit such attacks or block legitimate requests.

Empirical work reveals further limitations. Monitoring chains of thought can help detect hidden misbehavior, but training against such monitors can encourage concealment \citep{baker2025monitoring,korbak2025chain}. Moreover, \citet{kale2026reliable} show that monitoring becomes less reliable when agents know they are being observed, while \citet{motwani2024secret} show that agents can hide coordination through steganographic communication. Recent audits of deployed monitoring systems also identify gaps in coverage and opportunities for evasion \citep{rein2026redteaming}. Lastly, pre-deployment evaluation faces similar problems. \citet{deckenbach2026models} show that models fine-tuned on texts describing evaluation procedures refuse harmful prompts more often on safety benchmarks which suggests that evaluation awareness can inflate measured safety.

\paragraph{\textcolor{seaborndarkred}{(C2c)} Training for Alignment Optimizes Imperfect Proxies.}
Instead of verifying a finished model or filtering its outputs, one can try to build an aligned system \emph{by construction} during training. Modern approaches include RLHF \citep{ouyang2022training}, DPO \citep{rafailov2023direct}, and Constitutional AI \citep{bai2022constitutional}. In all cases, the model must generalize from finite preferences, examples, or principles to situations not observed during training.

A first problem is \emph{alignment misspecification}. Reward models, preference data, or written principles shift the alignment problem to the specification itself. These proxies may themselves be misaligned with the intended human objective and fail to capture it accurately. In reward-based methods, this can lead to \emph{reward hacking}, while \emph{goal misgeneralization} describes cases where a model behaves competently in a new setting but pursues the wrong goal \citep{di2022goal,ngo2024alignment}.

This is closely related to Goodhart's law, where optimization can break the relationship between a proxy and the underlying goal \citep{manheim2018categorizing}. \citet{skalse2022defining} show formally that optimizing a proxy reward can reduce the true reward unless strong conditions hold, and \citet{gao2023scaling} demonstrate this empirically as stronger optimization first improves and then reduces gold-standard performance. The problem is not specific to explicit reward models. DPO can also be misspecified when its implicit objective fails to represent the true preference structure \citep{gopalan2026dpo}, while \citet{khalaf2026inference} find related failures even under inference-time reward optimization with best-of-$n$ sampling.

A second problem is that aligned behavior does not necessarily imply robust alignment. \citet{ngo2024alignment} describe \emph{deceptive alignment}, where a model behaves as intended during training while preserving a different internal objective. \citet{greenblatt2024alignment} demonstrate alignment faking,
where a model behaves differently depending on whether it
believes its responses will be used for training. \citet{baum2026rewarded} similarly argue that reward-based training may establish conditional rather than unconditional compliance, and \citet{hubinger2024sleeper} show that deceptive strategies can survive supervised fine-tuning, reinforcement learning, and adversarial training. \citet{dev2026open} further find that safety training in open-weight models can be removed relatively easily compared with increasing their underlying biological capabilities.

Overall, training can strongly steer model behavior, but finite feedback neither guarantees that the intended objective is learned nor that apparently aligned behavior will persist outside the conditions in which it was trained and evaluated \citep{casper2023open,anwar2024foundational}.

\paragraph{\textcolor{seaborndarkred}{(C2d)} Useful and Harmful Capabilities Can Overlap.}
A further difficulty is that useful and harmful capabilities can rely on the same underlying knowledge. A model with strong scientific or engineering knowledge may therefore be able to infer harmful applications even if explicit instructions are suppressed. \citet{glukhov2025breach} demonstrate a related problem.
Individually safe responses can be combined to reveal harmful
information.

Existing work points to related trade-offs. \citet{wolf2023fundamental} show that behavior with nonzero probability can be elicited by sufficiently long prompts within their framework, while \citet{zhao2026gains} find a trade-off between removing harmful capabilities, avoiding excessive refusal, and preserving general utility. \citet{west2025ai} go further and argue that better alignment can make misalignment easier. A model that separates good from bad more sharply also makes it easier for an adversary to invert this distinction. \citet{mattsson2026fundamental} show in a game-theoretic model that if an AI cannot tell benign from malicious users, any advice that fully helps the benign user also helps the malicious one.

\begin{exampleboxred}{Cybersecurity}
Consider a model asked to build a secure chat application while being prohibited from attacking software. Suppose it recognizes that a widely used cryptographic library contains an unknown vulnerability and therefore avoids a particular function through an unusual workaround. The model never exploits the vulnerability, but the workaround itself may reveal where the flaw lies. Preventing such disclosures without also limiting the model's ability to reason about secure implementations may therefore be difficult.
\end{exampleboxred}

\paragraph{\textcolor{seaborndarkred}{(C2e)} Tail Risks Dominate at Scale.}
Finally, we argue that near-perfect alignment may not be good enough. One reason is repeated exposure. Even a very small failure rate can lead to many failures when a system is used millions of times.
A second reason is amplification. Autonomous replication can allow a single failure or deviation to affect many additional systems or interactions \citep{black2025replibench}.
A third reason is change over time. Competitive and evolutionary pressures may favor behaviors that improve persistence or performance at the expense of alignment \citep{hendrycks2023natural}. Experiments with self-evolving agents further suggest that such deviations can spread over repeated adaptation \citep{han2025alignment}.

We therefore cannot assume that making individual failures sufficiently rare makes the alignment problem practically solved.

\paragraph{Summary.}
We have argued that alignment cannot generally be guaranteed across relevant deployment conditions. This claim is supported both by empirical evidence of failure in current systems and by theoretical results that establish fundamental limits.
Verification, runtime guardrails, and alignment during training each leave important failure modes open. While worst-case analysis does not rule out success in restricted settings, it is appropriate when asking for general guarantees. Such guarantees must also hold in rare or difficult cases, not just under typical conditions.

There are important counterpoints to this claim. \emph{Guaranteed Safe AI} proposes combining a world model, safety specification, and verifier to provide quantitative safety guarantees \citep{dalrymple2024towards}. This offers a different route to safety, but its guarantees remain conditional on the chosen specification and world model. Assuming a sufficiently accurate world model is itself a strong assumption, especially in open and changing environments.

A second objection is that perfect alignment may simply be unnecessary if approximate alignment is good enough in practice. We agree that many applications do not require perfect behavior. However, as discussed in the last paragraph, rare failures can accumulate at scale, spread through replication or interaction, and be amplified by competitive pressures.

\subsection{\textcolor{seaborndarkviolet}{Claim 3}: Alignment is not sufficient}
Suppose that an individual AI system satisfies an agreed-upon alignment specification and is considered aligned in some reasonable sense. We now ask whether this model-level alignment is sufficient once such systems are deployed in human environments.

Alignment is established during training or enforced through mechanisms such as guardrails, but deployment takes place in an open world. There, AI systems must generalize beyond the conditions in which they were aligned and may cooperate or compete with other agents at large scale.

\paragraph{\textcolor{seaborndarkviolet}{(C3a)} AI Systems Change Their Environments.}
Human society is a complex system shaped by nonlinear relationships, feedback loops, and tipping points. Predicting the effects of interventions in such systems is difficult. More broadly, technical components cannot be understood independently of the social systems in which they are embedded \citep{selbst2019fairness}. Alignment based on consequences therefore requires more than understanding the model and human intentions. It also requires anticipating how an action changes the world around it.

Even relatively simple decision systems can create feedback loops by changing the data they later learn from. 

\begin{exampleboxviolet}{Runaway Feedback Loops}
\citet{ensign2018runaway} study predictive policing, where sending more police to an area produces more observed crime there. These observations are then used to update the model and can lead to still more policing of the same area. They show that this process can produce a runaway feedback loop even when observed crime accurately reflects the underlying crime rate wherever police are deployed.
\end{exampleboxviolet}

We argue that this is a problem for alignment because even an action that appears acceptable in isolation can change the environment and trigger feedback that amplifies its effects beyond what was intended.

\paragraph{\textcolor{seaborndarkviolet}{(C3b)} Small Changes Can Trigger Large Effects.}
Threshold effects are common in complex systems such as societies. Small changes can push a system across a threshold and produce qualitatively different outcomes. One relevant case for alignment is \emph{capability uplift}. Even a modest improvement in what an actor can do may enable actions that were previously out of reach.

\begin{exampleboxviolet}{Unfamiliar Cyber Target}
Consider a small hacking group that can exploit familiar systems but encounters a target built on an unfamiliar software stack. Without additional expertise, the group may abandon the attack. An aligned AI system can still provide individually permissible assistance such as explaining the architecture, interpreting logs, or adapting existing code to the new environment. Together, these capabilities may make the unfamiliar target tractable and substantially increase the probability of a successful intrusion.
\end{exampleboxviolet}

The problem is therefore not that the model directly performs a prohibited action, but that individually acceptable assistance can remove a critical bottleneck and push a harmful activity across a feasibility threshold \citep{anthropic2026misuse}.


AI can also create harm by making existing actions much cheaper and easier to scale. Many institutions rely on practical friction because large-scale action normally requires substantial time, money, or labor. Submitting hundreds of comments, applications, or filings, for example, was always possible but costly. AI can make such actions cheap enough to overwhelm systems built for human-scale input \citep{piedrahitaposition}. 

\paragraph{\textcolor{seaborndarkviolet}{(C3c)} Aligned Agents Can Produce Misaligned Collectives.}
Individual alignment does not guarantee aligned behavior at the group level. \citet{de2026conformity} study opinion dynamics and define an agent as aligned when it follows its own default position. They show that conformity can drive populations toward the opposite position. \citet{shen2026ai} compare individual and multi-agent systems built from models trained to follow Claude's constitution. They measure alignment by how well the resulting solutions follow its ethical principles. They find that multi-agent organizations achieve business goals more effectively but perform worse on ethical criteria. \citet{nakamura2026colosseum} give agents a shared cooperative objective. Even when agents are not instructed to collude, private communication can lead them to pursue conflicting goals and reduce collective performance. However, collusive discussions do not always lead to collusive actions. \citet{bajaj2026position} show that changing the order of interacting agents can substantially alter collective decisions, even when models and prompts remain identical.

These studies use different notions of alignment, and none establishes that the individual agents were perfectly aligned before interacting. Nevertheless, they show that safety training, individual preferences, and shared objectives do not necessarily produce desirable collective behavior.

\paragraph{\textcolor{seaborndarkviolet}{(C3d)} Collective Alignment Shifts the Problem.}
One response is to move from individual to collective alignment. Recent work distinguishes \emph{intra-agent alignment} from \emph{collective alignment}, which asks whether interacting agents remain aligned as a group \citep{hung2026deliberation}. We agree that this is an important extension.

However, the need for collective alignment reveals a deeper problem. Alignment does not compose. If model-level alignment is supposed to justify safe deployment, then it must survive interaction with other aligned systems. If a separate notion of group alignment is required, that guarantee has not carried over from the model to the collective. The same problem then reappears when collectives interact with humans, institutions, and changing incentives. Alignment therefore has to be re-established at each level of organization rather than carrying over from the parts to the whole.

\paragraph{Summary.}
Taken together, these arguments support our claim that model-level alignment is not sufficient for societal safety. Even aligned systems can cause harm once they interact with other agents, institutions, and changing social environments.

A different view is that AI should be treated as a \emph{normal technology} whose effects remain mediated by people and institutions \citep{narayanan2025ai}. We agree, but argue that institutions themselves adapt to new capabilities and incentives, which can create new systemic risks.
Another response is to design agents that avoid harmful side effects. \citet{krakovna2020avoiding}, for example, propose preserving the ability to perform future tasks. This can reduce side effects in controlled settings, but in open social environments it is much harder to implement. 

One might also object that alignment is meant to prevent catastrophic risks, not every negative effect of AI. However, this assumes that catastrophic risks can be identified in advance. In a complex society, small changes can scale, interact, and cross thresholds in ways that are difficult to foresee.

\section{Alternative View: Emergent Risk in Resilient Hybrid Societies}

Given the limits of what alignment can capture, what can be solved, and the often unforeseeable pathways through which harm can arise, we argue for a more pragmatic view of mitigating AI risk.

The definition proposed by \citet{mckinlay2026understanding} is closer to the view we advocate. They define value alignment as
\begin{quote}
``an ongoing process between humans and autonomous agents that aims to express and implement abstract values in diverse contexts, while managing the cognitive limits of both humans and AI agents and also balancing the conflicting ethical and political demands generated by the values in different groups.''
\end{quote}

A key task is therefore to understand the pathways through which AI can cause harm, especially through feedback effects and indirect consequences. We argue that alignment alone cannot prevent these harms. Small shifts in capability can already have large effects. This suggests placing greater emphasis on assessing potential societal effects during alignment training and monitoring actual effects after deployment. Such assessments should consider not only individual model behavior, but also interactions between AI systems and their effects on human institutions.  We agree with \citet{perez-ortiz2026position} that such assessments should also consider long-term effects on ecological systems.

\citet{carichon2025coming} call for simulation benchmarks that examine how cooperation and competition between AI agents can undermine individual and collective alignment. For AI agents sharing resources such as compute or memory, \citet{truong2026multi} propose evaluating whether institutional rules, monitoring, sanctions, and repair mechanisms prevent resource overuse and enable recovery from failures. When harmful effects emerge, AI developers and affected institutions should be prepared to respond quickly and adapt together. For instance, if AI agents begin overwhelming public submission systems, developers and institutions should directly coordinate to mitigate the problem.

Human law, morality, and institutions developed in a non-AI world and will need to adapt to hybrid societies. \citet{gabriel2025we} argue that a world of AI agents creates new challenges for human--machine relationships and social coordination. Human societies have undergone major transitions before, from small groups based largely on direct relationships to societies of millions coordinated through institutions and shared rules. The internet created another such shift by enabling near-instant global communication and changing both how society functions and where it is vulnerable.

Society must therefore become more resilient to AI-related harm, including harms that emerge from interactions among AI systems, people, and institutions, as well as cross-border effects from systems legal in one country but illegal in another.
Assigning responsibility also remains crucial. Delegating actions to AI should not erase human or institutional responsibility. Those who deploy AI agents, enable their actions, or provide them with resources should remain identifiable and accountable for those choices and effects.

AI adoption and risk management should therefore be treated as an ongoing process. This requires continuously monitoring how AI systems affect society rather than relying only on one-time certification, and repeatedly reassessing which uses require restriction or adaptation. Resilience must also be built fairly, considering not only where AI creates benefits but also where its harms are distributed.

\section{Conclusions}

We have argued that human-value alignment is fundamentally underspecified, that general guarantees remain unavailable even when a target can be specified, and that model-level alignment would still not be sufficient for societal safety.
This does not mean that alignment research lacks value. Much of it provides useful insight into how models fail.

Our contribution goes beyond identifying limitations of individual alignment. We argue that human intentions may be ambiguous even at deployment time, while judging alignment by consequences requires a sufficiently accurate world model, which is practically infeasible in open environments. More importantly, alignment does not generally compose across levels of organization. Even agents trained for alignment can produce harmful collective outcomes, while rare failures can accumulate and amplify at societal scale. Through \emph{performative target shift}, AI systems may even change the preferences and values against which they are evaluated. To the best of our knowledge, previous work has not brought these limitations together to show why alignment cannot guarantee safety in hybrid human--AI societies.

Related work points in a similar direction. \citet{mckinlay2026understanding} define value alignment as an ongoing process between humans and autonomous agents rather than a fixed property. \citet{han2026social} argue that humans and AI increasingly co-evolve within hybrid societies. Both suggest that safety requires continuous observation and adaptation, not a one-time technical solution.

Our main message is therefore one of caution. \textbf{Waiting for a solution to the alignment problem that may never come risks delaying measures that are already needed.} Resources should be invested now in mitigating AI harm and making institutions and societies more resilient to the disruptions and risks created by AI.

\subsection*{AI use statement}

We used generative AI tools to improve wording and readability and to help identify potentially relevant literature. All suggestions were reviewed by the authors, who take full responsibility for the final content.

\subsection*{Ethics statement}

AI alignment and AI safety research may substantially influence how resources are allocated to reduce risks from increasingly capable AI systems. These choices will have significant consequences for human well-being. We consider it important to scrutinize the assumptions underlying alignment research as well as the opportunity costs of  particular technical approaches.

\bibliography{iclr2027_conference}

\begin{thebibliography}{94}
\providecommand{\natexlab}[1]{#1}
\providecommand{\url}[1]{\texttt{#1}}
\expandafter\ifx\csname urlstyle\endcsname\relax
  \providecommand{\doi}[1]{doi: #1}\else
  \providecommand{\doi}{doi: \begingroup \urlstyle{rm}\Url}\fi

\bibitem[Abdelnabi \& Bagdasarian(2026)Abdelnabi and Bagdasarian]{abdelnabi2026ai}
Sahar Abdelnabi and Eugene Bagdasarian.
\newblock {AI} agents may always fall for prompt injections.
\newblock \emph{arXiv preprint arXiv:2605.17634}, 2026.

\bibitem[Agarwal(2026)]{agarwal2026no}
Ayushi Agarwal.
\newblock No certificate for alignment: Two independent impossibilities and the {Pareto} frontier of achievable safety guarantees.
\newblock \emph{arXiv preprint arXiv:2603.08761}, 2026.

\bibitem[Alfonseca et~al.(2021)Alfonseca, Cebrian, Anta, Coviello, Abeliuk, and Rahwan]{alfonseca2021superintelligence}
Manuel Alfonseca, Manuel Cebrian, Antonio~Fernandez Anta, Lorenzo Coviello, Andr{\'e}s Abeliuk, and Iyad Rahwan.
\newblock Superintelligence cannot be contained: Lessons from computability theory.
\newblock \emph{Journal of Artificial Intelligence Research}, 70:\penalty0 65--76, 2021.

\bibitem[{Anthropic}(2026{\natexlab{a}})]{anthropic2026misuse}
{Anthropic}.
\newblock Detecting and countering misuse of {AI}: {September} 2026.
\newblock \url{https://www.anthropic.com/threat-intelligence-report-september-2026}, September 2026{\natexlab{a}}.
\newblock Published September 10, 2026.

\bibitem[{Anthropic}(2026{\natexlab{b}})]{anthropic2026multiagent}
{Anthropic}.
\newblock Multi-agent systems: Research and perspectives, January 2026{\natexlab{b}}.
\newblock URL \url{https://www.anthropic.com/research/multiagent-systems}.
\newblock Accessed: 2026-09-09.

\bibitem[Anwar et~al.(2024)Anwar, Saparov, Rando, Paleka, Turpin, Hase, Lubana, Jenner, Casper, Sourbut, et~al.]{anwar2024foundational}
Usman Anwar, Abulhair Saparov, Javier Rando, Daniel Paleka, Miles Turpin, Peter Hase, Ekdeep~Singh Lubana, Erik Jenner, Stephen Casper, Oliver Sourbut, et~al.
\newblock Foundational challenges in assuring alignment and safety of large language models.
\newblock \emph{arXiv preprint arXiv:2404.09932}, 2024.

\bibitem[Bai et~al.(2022)Bai, Kadavath, Kundu, Askell, Kernion, Jones, Chen, Goldie, Mirhoseini, McKinnon, et~al.]{bai2022constitutional}
Yuntao Bai, Saurav Kadavath, Sandipan Kundu, Amanda Askell, Jackson Kernion, Andy Jones, Anna Chen, Anna Goldie, Azalia Mirhoseini, Cameron McKinnon, et~al.
\newblock Constitutional {AI}: Harmlessness from {AI} feedback.
\newblock \emph{arXiv preprint arXiv:2212.08073}, 2022.

\bibitem[Bajaj et~al.(2026)Bajaj, Singh, Anand, and Singh]{bajaj2026position}
Tanav~Singh Bajaj, Nikhil Singh, Karan Anand, and Eishkaran Singh.
\newblock Position: Safety and fairness in agentic {AI} depend on interaction topology, not on model scale or alignment.
\newblock \emph{arXiv preprint arXiv:2605.01147}, 2026.

\bibitem[Baker et~al.(2025)Baker, Huizinga, Gao, Dou, Guan, Madry, Zaremba, Pachocki, and Farhi]{baker2025monitoring}
Bowen Baker, Joost Huizinga, Leo Gao, Zehao Dou, Melody~Y Guan, Aleksander Madry, Wojciech Zaremba, Jakub Pachocki, and David Farhi.
\newblock Monitoring reasoning models for misbehavior and the risks of promoting obfuscation.
\newblock \emph{arXiv preprint arXiv:2503.11926}, 2025.

\bibitem[Ball \& Hackemann(2026)Ball and Hackemann]{ball2026position}
Sarah Ball and Phil Hackemann.
\newblock Position: The alignment community is unintentionally building a censor's toolkit.
\newblock \emph{arXiv preprint arXiv:2608.12346}, 2026.

\bibitem[Ball et~al.(2026)Ball, Gluch, Goldwasser, Kreuter, Reingold, and Rothblum]{ball2026impossibility}
Sarah Ball, Greg Gluch, Shafi Goldwasser, Frauke Kreuter, Omer Reingold, and Guy Rothblum.
\newblock On the impossibility of separating intelligence from judgment: The computational intractability of filtering for {AI} alignment.
\newblock In \emph{International Conference on Learning Representations}, volume 2026, pp.\  65502--65534, 2026.

\bibitem[Bar-Shalom et~al.(2026)Bar-Shalom, Geva, Frasca, and Maron]{bar2026lemon}
Guy Bar-Shalom, Mor Geva, Fabrizio Frasca, and Haggai Maron.
\newblock {LEMON}: Equivariant learning over attention maps for zero-shot misbehavior monitoring on new {LLMs}.
\newblock \emph{arXiv preprint arXiv:2610.22723}, 2026.

\bibitem[Baum(2024)]{baum2024disentangling}
Kevin Baum.
\newblock Disentangling {AI} alignment: a structured taxonomy beyond safety and ethics.
\newblock In \emph{International Conference on Bridging the Gap between AI and Reality}, pp.\  158--173. Springer, 2024.

\bibitem[Baum \& Slavkovik(2025)Baum and Slavkovik]{baum2025aggregation}
Kevin Baum and Marija Slavkovik.
\newblock Aggregation problems in machine ethics and {AI} alignment.
\newblock In \emph{Proceedings of the AAAI/ACM Conference on AI, Ethics, and Society}, volume~8, pp.\  355--366, 2025.

\bibitem[Baum et~al.(2026)Baum, Binkyt{\.e}, Gro{\ss}mann, Jahn, and Mannino]{baum2026rewarded}
Kevin Baum, Ruta Binkyt{\.e}, Gerrit Gro{\ss}mann, Felix Jahn, and Adriano Mannino.
\newblock Rewarded for not getting caught: Why alignment training selects for conditional compliance and cannot instill norms.
\newblock \emph{arXiv preprint arXiv:2611.02162}, 2026.

\bibitem[Bengio et~al.(2023)Bengio, Hinton, Yao, Song, Abbeel, Harari, Zhang, Xue, Shalev-Shwartz, Hadfield, et~al.]{bengio2023managing}
Yoshua Bengio, Geoffrey Hinton, Andrew Yao, Dawn Song, Pieter Abbeel, Yuval~Noah Harari, Ya-Qin Zhang, Lan Xue, Shai Shalev-Shwartz, Gillian Hadfield, et~al.
\newblock Managing {AI} risks in an era of rapid progress.
\newblock \emph{arXiv preprint arXiv:2310.17688}, pp.\ ~18, 2023.

\bibitem[Bergstr{\"a}{\ss}er et~al.(2026)Bergstr{\"a}{\ss}er, Cotterell, and Lin]{bergstrasser2026transformers}
Pascal Bergstr{\"a}{\ss}er, Ryan Cotterell, and Anthony~W Lin.
\newblock Transformers are inherently succinct.
\newblock In \emph{International Conference on Learning Representations}, volume 2026, pp.\  58169--58189, 2026.

\bibitem[Black et~al.(2025)Black, Stickland, Pencharz, Sourbut, Schmatz, Bailey, Matthews, Millwood, Remedios, and Cooney]{black2025replibench}
Sid Black, Asa~Cooper Stickland, Jake Pencharz, Oliver Sourbut, Michael Schmatz, Jay Bailey, Ollie Matthews, Ben Millwood, Alex Remedios, and Alan Cooney.
\newblock Replibench: Evaluating the autonomous replication capabilities of language model agents.
\newblock \emph{arXiv preprint arXiv:2504.18565}, 2025.

\bibitem[Bolukbasi et~al.(2016)Bolukbasi, Chang, Zou, Saligrama, and Kalai]{bolukbasi2016man}
Tolga Bolukbasi, Kai-Wei Chang, James~Y Zou, Venkatesh Saligrama, and Adam~T Kalai.
\newblock Man is to computer programmer as woman is to homemaker? debiasing word embeddings.
\newblock \emph{Advances in neural information processing systems}, 29, 2016.

\bibitem[Carichon et~al.(2025)Carichon, Khandelwal, Fauchard, and Farnadi]{carichon2025coming}
Florian Carichon, Aditi Khandelwal, Marylou Fauchard, and Golnoosh Farnadi.
\newblock The coming crisis of multi-agent misalignment: {AI} alignment must be a dynamic and social process.
\newblock \emph{arXiv preprint arXiv:2506.01080}, 2025.

\bibitem[Carroll et~al.(2024)Carroll, Foote, Siththaranjan, Russell, and Dragan]{carroll2024ai}
Micah Carroll, Davis Foote, Anand Siththaranjan, Stuart Russell, and Anca Dragan.
\newblock {AI} alignment with changing and influenceable reward functions.
\newblock \emph{arXiv preprint arXiv:2405.17713}, 2024.

\bibitem[Casper et~al.(2023)Casper, Davies, Shi, Gilbert, Scheurer, Rando, Freedman, Korbak, Lindner, Freire, et~al.]{casper2023open}
Stephen Casper, Xander Davies, Claudia Shi, Thomas~Krendl Gilbert, J{\'e}r{\'e}my Scheurer, Javier Rando, Rachel Freedman, Tomasz Korbak, David Lindner, Pedro Freire, et~al.
\newblock Open problems and fundamental limitations of reinforcement learning from human feedback.
\newblock \emph{arXiv preprint arXiv:2307.15217}, 2023.

\bibitem[Cunningham et~al.(2026)Cunningham, Wei, Wang, Persic, Peng, Abderrachid, Agarwal, Chen, Cohen, Dau, et~al.]{cunningham2026constitutional}
Hoagy Cunningham, Jerry Wei, Zihan Wang, Andrew Persic, Alwin Peng, Jordan Abderrachid, Raj Agarwal, Bobby Chen, Austin Cohen, Andy Dau, et~al.
\newblock Constitutional classifiers++: Efficient production-grade defenses against universal jailbreaks.
\newblock \emph{arXiv preprint arXiv:2601.04603}, 2026.

\bibitem[Dafoe et~al.(2020)Dafoe, Hughes, Bachrach, Collins, McKee, Leibo, Larson, and Graepel]{dafoe2020open}
Allan Dafoe, Edward Hughes, Yoram Bachrach, Tantum Collins, Kevin~R McKee, Joel~Z Leibo, Kate Larson, and Thore Graepel.
\newblock Open problems in cooperative {AI}.
\newblock \emph{arXiv preprint arXiv:2012.08630}, 2020.

\bibitem[Dafoe et~al.(2021)Dafoe, Bachrach, Hadfield, Horvitz, Larson, and Graepel]{dafoe2021cooperative}
Allan Dafoe, Yoram Bachrach, Gillian Hadfield, Eric Horvitz, Kate Larson, and Thore Graepel.
\newblock Cooperative {AI}: machines must learn to find common ground.
\newblock \emph{Nature}, 593\penalty0 (7857):\penalty0 33--36, 2021.

\bibitem[Dahlgren~Lindstr{\"o}m et~al.(2025)Dahlgren~Lindstr{\"o}m, Methnani, Krause, Ericson, de~Rituerto~de Troya, Coelho~Mollo, and Dobbe]{dahlgren2025helpful}
Adam Dahlgren~Lindstr{\"o}m, Leila Methnani, Lea Krause, Petter Ericson, {\'I}{\~n}igo~Mart{\'\i}nez de~Rituerto~de Troya, Dimitri Coelho~Mollo, and Roel Dobbe.
\newblock Helpful, harmless, honest? sociotechnical limits of {AI} alignment and safety through reinforcement learning from human feedback: Ad lindstr{\"o}m et al.
\newblock \emph{Ethics and Information Technology}, 27\penalty0 (2):\penalty0 28, 2025.

\bibitem[Dai \& Xiao(2025)Dai and Xiao]{dai2026embracing}
Gordon Dai and Yunze Xiao.
\newblock Embracing contradiction: Theoretical inconsistency will not impede the road of building responsible {AI} systems.
\newblock \emph{Advances in Neural Information Processing Systems}, 38, 2025.

\bibitem[Dalrymple et~al.(2024)Dalrymple, Skalse, Bengio, Russell, Tegmark, Seshia, Omohundro, Szegedy, Goldhaber, Ammann, et~al.]{dalrymple2024towards}
David Dalrymple, Joar Skalse, Yoshua Bengio, Stuart Russell, Max Tegmark, Sanjit Seshia, Steve Omohundro, Christian Szegedy, Ben Goldhaber, Nora Ammann, et~al.
\newblock Towards guaranteed safe {AI}: A framework for ensuring robust and reliable {AI} systems.
\newblock \emph{arXiv preprint arXiv:2405.06624}, 2024.

\bibitem[De~Marzo et~al.(2026)De~Marzo, Bellina, Castellano, Priesemann, and Garcia]{de2026conformity}
Giordano De~Marzo, Alessandro Bellina, Claudio Castellano, Viola Priesemann, and David Garcia.
\newblock Conformity generates collective misalignment in {AI} agents societies.
\newblock \emph{arXiv preprint arXiv:2605.10721}, 2026.

\bibitem[Deckenbach et~al.(2026)Deckenbach, Puerto, Geiping, and Abdelnabi]{deckenbach2026models}
Katharina Deckenbach, Haritz Puerto, Jonas Geiping, and Sahar Abdelnabi.
\newblock Models that know how evaluations are designed score safer.
\newblock \emph{arXiv preprint arXiv:2605.28591}, 2026.

\bibitem[Delobelle et~al.(2022)Delobelle, Tokpo, Calders, and Berendt]{delobelle2022measuring}
Pieter Delobelle, Ewoenam Tokpo, Toon Calders, and Bettina Berendt.
\newblock Measuring fairness with biased rulers: A comparative study on bias metrics for pre-trained language models.
\newblock In \emph{Proceedings of the 2022 Conference of the North American Chapter of the Association for Computational Linguistics: Human Language Technologies}, pp.\  1693--1706, 2022.

\bibitem[Dev et~al.(2026)Dev, Barkan, Hakim, Hollister, Lee, Worland, Maciorowski, and Guerra]{dev2026open}
Sunishchal Dev, Casey~O. Barkan, Anthony Hakim, Brooke Hollister, Jeffrey Lee, Alyssa Worland, Dawid Maciorowski, and Steph Guerra.
\newblock Open-weight {AI} models may increase biological misuse risks.
\newblock 2026.

\bibitem[Di~Langosco et~al.(2022)Di~Langosco, Koch, Sharkey, Pfau, and Krueger]{di2022goal}
Lauro~Langosco Di~Langosco, Jack Koch, Lee~D Sharkey, Jacob Pfau, and David Krueger.
\newblock Goal misgeneralization in deep reinforcement learning.
\newblock In \emph{International Conference on Machine Learning}, pp.\  12004--12019. PMLR, 2022.

\bibitem[Eckersley(2018)]{eckersley2018impossibility}
Peter Eckersley.
\newblock Impossibility and uncertainty theorems in {AI} value alignment (or why your {AGI} should not have a utility function).
\newblock \emph{arXiv preprint arXiv:1901.00064}, 2018.

\bibitem[Ensign et~al.(2018)Ensign, Friedler, Neville, Scheidegger, and Venkatasubramanian]{ensign2018runaway}
Danielle Ensign, Sorelle~A Friedler, Scott Neville, Carlos Scheidegger, and Suresh Venkatasubramanian.
\newblock Runaway feedback loops in predictive policing.
\newblock In \emph{Conference on fairness, accountability and transparency}, pp.\  160--171. PMLR, 2018.

\bibitem[Gabriel(2020)]{gabriel2020artificial}
Iason Gabriel.
\newblock Artificial intelligence, values, and alignment: I. gabriel.
\newblock \emph{Minds and machines}, 30\penalty0 (3):\penalty0 411--437, 2020.

\bibitem[Gabriel \& Keeling(2025)Gabriel and Keeling]{gabriel2025matter}
Iason Gabriel and Geoff Keeling.
\newblock A matter of principle? {AI} alignment as the fair treatment of claims.
\newblock \emph{Philosophical Studies}, 182\penalty0 (7):\penalty0 1951--1973, 2025.

\bibitem[Gabriel et~al.(2025)Gabriel, Keeling, Manzini, and Evans]{gabriel2025we}
Iason Gabriel, Geoff Keeling, Arianna Manzini, and James Evans.
\newblock We need a new ethics for a world of {AI} agents.
\newblock \emph{Nature}, 644\penalty0 (8075):\penalty0 38--40, 2025.

\bibitem[Gao et~al.(2023)Gao, Schulman, and Hilton]{gao2023scaling}
Leo Gao, John Schulman, and Jacob Hilton.
\newblock Scaling laws for reward model overoptimization.
\newblock In \emph{International conference on machine learning}, pp.\  10835--10866. PMLR, 2023.

\bibitem[Ge et~al.(2024)Ge, Halpern, Micha, Procaccia, Shapira, Vorobeychik, and Wu]{ge2024axioms}
Luise Ge, Daniel Halpern, Evi Micha, Ariel~D Procaccia, Itai Shapira, Yevgeniy Vorobeychik, and Junlin Wu.
\newblock Axioms for {AI} alignment from human feedback.
\newblock \emph{Advances in Neural Information Processing Systems}, 37:\penalty0 80439--80465, 2024.

\bibitem[Glukhov et~al.(2025)Glukhov, Han, Shumailov, Papyan, and Papernot]{glukhov2025breach}
David Glukhov, Ziwen Han, Ilia Shumailov, Vardan Papyan, and Nicolas Papernot.
\newblock Breach by a thousand leaks: Unsafe information leakage in'safe'{AI} responses.
\newblock In \emph{International Conference on Learning Representations}, volume 2025, pp.\  51569--51600, 2025.

\bibitem[Gonen \& Goldberg(2019)Gonen and Goldberg]{gonen2019lipstick}
Hila Gonen and Yoav Goldberg.
\newblock Lipstick on a pig: Debiasing methods cover up systematic gender biases in word embeddings but do not remove them.
\newblock In \emph{Proceedings of the 2019 Conference of the North American Chapter of the Association for Computational Linguistics: Human Language Technologies, Volume 1 (Long and Short Papers)}, pp.\  609--614, 2019.

\bibitem[Gopalan et~al.(2026)Gopalan, Ray~Chowdhury, and Banerjee]{gopalan2026dpo}
Aditya Gopalan, Sayak Ray~Chowdhury, and Debangshu Banerjee.
\newblock Why {DPO} is a misspecified estimator and how to fix it.
\newblock In \emph{International Conference on Learning Representations}, volume 2026, pp.\  88837--88847, 2026.

\bibitem[Greenblatt et~al.(2023)Greenblatt, Shlegeris, Sachan, and Roger]{greenblatt2023ai}
Ryan Greenblatt, Buck Shlegeris, Kshitij Sachan, and Fabien Roger.
\newblock {AI} control: Improving safety despite intentional subversion.
\newblock \emph{arXiv preprint arXiv:2312.06942}, 2023.

\bibitem[Greenblatt et~al.(2024)Greenblatt, Denison, Wright, Roger, MacDiarmid, Marks, Treutlein, Belonax, Chen, Duvenaud, et~al.]{greenblatt2024alignment}
Ryan Greenblatt, Carson Denison, Benjamin Wright, Fabien Roger, Monte MacDiarmid, Sam Marks, Johannes Treutlein, Tim Belonax, Jack Chen, David Duvenaud, et~al.
\newblock Alignment faking in large language models.
\newblock \emph{arXiv preprint arXiv:2412.14093}, 2\penalty0 (1), 2024.

\bibitem[Guzman~Piedrahita et~al.(2026)Guzman~Piedrahita, Li, Banerjee, Zhang, Blin, Simko, Pandey, Strauss, Mihalcea, Sch{\"o}lkopf, et~al.]{piedrahitaposition}
David Guzman~Piedrahita, Changling Li, Dave Banerjee, Terry~Jingchen Zhang, Kevin Blin, Samuel Simko, Punya~Syon Pandey, Irene Strauss, Rada Mihalcea, Bernhard Sch{\"o}lkopf, et~al.
\newblock Position: Safe models do not guarantee safe societies.
\newblock In \emph{Forty-third International Conference on Machine Learning Position Paper Track}, 2026.

\bibitem[Hadfield-Menell et~al.(2016)Hadfield-Menell, Russell, Abbeel, and Dragan]{hadfield2016cooperative}
Dylan Hadfield-Menell, Stuart~J Russell, Pieter Abbeel, and Anca Dragan.
\newblock Cooperative inverse reinforcement learning.
\newblock \emph{Advances in neural information processing systems}, 29, 2016.

\bibitem[Hammond et~al.(2025)Hammond, Chan, Clifton, Hoelscher-Obermaier, Khan, McLean, Smith, Barfuss, Foerster, Gaven{\v{c}}iak, et~al.]{hammond2025multi}
Lewis Hammond, Alan Chan, Jesse Clifton, Jason Hoelscher-Obermaier, Akbir Khan, Euan McLean, Chandler Smith, Wolfram Barfuss, Jakob Foerster, Tom{\'a}{\v{s}} Gaven{\v{c}}iak, et~al.
\newblock Multi-agent risks from advanced {AI}.
\newblock \emph{arXiv preprint arXiv:2502.14143}, 2025.

\bibitem[Han et~al.(2025)Han, Xiong, Liu, Ye, Su, Duan, Liu, Xie, Bansal, Ding, et~al.]{han2025alignment}
Siwei Han, Kaiwen Xiong, Jiaqi Liu, Xinyu Ye, Yaofeng Su, Wenbo Duan, Xinyuan Liu, Cihang Xie, Mohit Bansal, Mingyu Ding, et~al.
\newblock Alignment tipping process: How self-evolution pushes {LLM} agents off the rails.
\newblock \emph{arXiv preprint arXiv:2510.04860}, 2025.

\bibitem[Han et~al.(2026)Han, Leibo, Lenaerts, Rahwan, Santos, Perc, and Capraro]{han2026social}
The~Anh Han, Joel~Z Leibo, Tom Lenaerts, Iyad Rahwan, Fernando Santos, Matja{\v{z}} Perc, and Valerio Capraro.
\newblock Social physics in the age of artificial intelligence.
\newblock \emph{arXiv preprint arXiv:2603.16900}, 2026.

\bibitem[Hendrycks(2023)]{hendrycks2023natural}
Dan Hendrycks.
\newblock Natural selection favors {AIs} over humans.
\newblock \emph{arXiv preprint arXiv:2303.16200}, 2023.

\bibitem[Hubinger et~al.(2024)Hubinger, Denison, Mu, Lambert, Tong, MacDiarmid, Lanham, Ziegler, Maxwell, Cheng, et~al.]{hubinger2024sleeper}
Evan Hubinger, Carson Denison, Jesse Mu, Mike Lambert, Meg Tong, Monte MacDiarmid, Tamera Lanham, Daniel~M Ziegler, Tim Maxwell, Newton Cheng, et~al.
\newblock Sleeper agents: Training deceptive {LLMs} that persist through safety training.
\newblock \emph{arXiv preprint arXiv:2401.05566}, 2024.

\bibitem[Hung et~al.(2026)Hung, Ward, Marx, Strickler, and Jensen]{hung2026deliberation}
Man Hung, Corban Ward, Jacob Marx, Isabella Strickler, and Annabella Jensen.
\newblock Deliberation and drift: Evaluating alignment fragility in multi-agent medical artificial intelligence.
\newblock \emph{AI and Ethics}, 6\penalty0 (2):\penalty0 198, 2026.

\bibitem[Jha et~al.(2026)Jha, {\v{S}}varc, and Bagi{\'n}ski]{jha2026position}
Tushita Jha, Rory {\v{S}}varc, and Mateusz Bagi{\'n}ski.
\newblock Position:'{AI} alignment'encompasses competing technical priorities.
\newblock In \emph{Forty-third International Conference on Machine Learning Position Paper Track}, 2026.

\bibitem[Jhala \& Majumdar(2009)Jhala and Majumdar]{jhala2009software}
Ranjit Jhala and Rupak Majumdar.
\newblock Software model checking.
\newblock \emph{ACM Computing Surveys (CSUR)}, 41\penalty0 (4):\penalty0 1--54, 2009.

\bibitem[Ji et~al.(2023)Ji, Qiu, Chen, Zhang, Lou, Wang, Duan, He, Vierling, Hong, et~al.]{ji2023ai}
Jiaming Ji, Tianyi Qiu, Boyuan Chen, Borong Zhang, Hantao Lou, Kaile Wang, Yawen Duan, Zhonghao He, Lukas Vierling, Donghai Hong, et~al.
\newblock {AI} alignment: A comprehensive survey.
\newblock \emph{arXiv preprint arXiv:2310.19852}, 2023.

\bibitem[Kale et~al.(2026)Kale, Zhang, Zhu, Aich, Rodriguez, Knight, and Wang]{kale2026reliable}
Neil Kale, Chen Bo~Calvin Zhang, Kevin Zhu, Ankit Aich, Paula Rodriguez, Christina Knight, and Zifan Wang.
\newblock Reliable weak-to-strong monitoring of {LLM} agents.
\newblock In \emph{International Conference on Learning Representations}, volume 2026, pp.\  78083--78138, 2026.

\bibitem[Kamin \& Rachlinski(1995)Kamin and Rachlinski]{kamin1995ex}
Kim~A Kamin and Jeffrey~J Rachlinski.
\newblock Ex post$\ne$ ex ante: Determining liability in hindsight.
\newblock \emph{Law and Human Behavior}, 19\penalty0 (1):\penalty0 89, 1995.

\bibitem[Katz et~al.(2017)Katz, Barrett, Dill, Julian, and Kochenderfer]{katz2017reluplex}
Guy Katz, Clark Barrett, David~L Dill, Kyle Julian, and Mykel~J Kochenderfer.
\newblock Reluplex: An efficient {SMT} solver for verifying deep neural networks.
\newblock In \emph{International conference on computer aided verification}, pp.\  97--117. Springer, 2017.

\bibitem[Keswani et~al.(2026{\natexlab{a}})Keswani, Cousins, Nguyen, Conitzer, Heidari, Borg, and Sinnott-Armstrong]{keswani2026moral}
Vijay Keswani, Cyrus Cousins, Breanna Nguyen, Vincent Conitzer, Hoda Heidari, Jana~Schaich Borg, and Walter Sinnott-Armstrong.
\newblock Moral change or noise? on problems of aligning {AI} with temporally unstable human feedback.
\newblock In \emph{Proceedings of the AAAI Conference on Artificial Intelligence}, volume~40, pp.\  37501--37509, 2026{\natexlab{a}}.

\bibitem[Keswani et~al.(2026{\natexlab{b}})Keswani, Nguyen, Cousins, Conitzer, Sinnott-Armstrong, and Borg]{keswani2026position}
Vijay Keswani, Breanna~K Nguyen, Cyrus Cousins, Vincent Conitzer, Walter Sinnott-Armstrong, and Jana~Schaich Borg.
\newblock Position: We need practical {AI} alignment methods to mirror human reasoning.
\newblock \emph{arXiv preprint arXiv:2608.12372}, 2026{\natexlab{b}}.

\bibitem[Khalaf et~al.(2025)Khalaf, Mayrink~Verdun, Oesterling, Lakkaraju, and Calmon]{khalaf2026inference}
Hadi Khalaf, Claudio Mayrink~Verdun, Alex Oesterling, Himabindu Lakkaraju, and Flavio Calmon.
\newblock Inference-time reward hacking in large language models.
\newblock \emph{Advances in Neural Information Processing Systems}, 38:\penalty0 61720--61760, 2025.

\bibitem[Korbak et~al.(2025)Korbak, Balesni, Barnes, Bengio, Benton, Bloom, Chen, Cooney, Dafoe, Dragan, et~al.]{korbak2025chain}
Tomek Korbak, Mikita Balesni, Elizabeth Barnes, Yoshua Bengio, Joe Benton, Joseph Bloom, Mark Chen, Alan Cooney, Allan Dafoe, Anca Dragan, et~al.
\newblock Chain of thought monitorability: A new and fragile opportunity for {AI} safety.
\newblock \emph{arXiv preprint arXiv:2507.11473}, 2025.

\bibitem[Krakovna et~al.(2020)Krakovna, Orseau, Ngo, Martic, and Legg]{krakovna2020avoiding}
Victoria Krakovna, Laurent Orseau, Richard Ngo, Miljan Martic, and Shane Legg.
\newblock Avoiding side effects by considering future tasks.
\newblock \emph{Advances in Neural Information Processing Systems}, 33:\penalty0 19064--19074, 2020.

\bibitem[Leike et~al.(2018)Leike, Krueger, Everitt, Martic, Maini, and Legg]{leike2018scalable}
Jan Leike, David Krueger, Tom Everitt, Miljan Martic, Vishal Maini, and Shane Legg.
\newblock Scalable agent alignment via reward modeling: a research direction.
\newblock \emph{arXiv preprint arXiv:1811.07871}, 2018.

\bibitem[Lindner et~al.(2025)Lindner, Griffin, Korbak, Zimmermann, Irving, Farquhar, and Cooney]{lindner2025practical}
David Lindner, Charlie Griffin, Tomek Korbak, Roland~S Zimmermann, Geoffrey Irving, Sebastian Farquhar, and Alan Cooney.
\newblock Practical challenges of control monitoring in frontier {AI} deployments.
\newblock \emph{arXiv preprint arXiv:2512.22154}, 2025.

\bibitem[{Machine Intelligence Research Institute}(2026)]{miriAbout}
{Machine Intelligence Research Institute}.
\newblock About {MIRI}.
\newblock \url{https://intelligence.org/about/}, 2026.
\newblock Accessed: 2026-09-09.

\bibitem[Manheim \& Garrabrant(2018)Manheim and Garrabrant]{manheim2018categorizing}
David Manheim and Scott Garrabrant.
\newblock Categorizing variants of {Goodhart}'s law.
\newblock \emph{arXiv preprint arXiv:1803.04585}, 2018.

\bibitem[Mattsson et~al.(2026)Mattsson, Berker, Tewolde, Anagnostides, Fragkia, Liu, Heckett, Liu, Hartman, Kova{\v{r}}{\'\i}k, et~al.]{mattsson2026fundamental}
Ida Mattsson, Ratip~Emin Berker, Emanuel Tewolde, Ioannis Anagnostides, Kiriaki Fragkia, Xinyu Liu, Alexander Heckett, Jiayuan Liu, Eden Hartman, Vojt{\v{e}}ch Kova{\v{r}}{\'\i}k, et~al.
\newblock Fundamental impossibility results in {AI} safety: Trading off between desiderata.
\newblock \emph{arXiv preprint arXiv:2610.27373}, 2026.

\bibitem[McKinlay et~al.(2026)McKinlay, De~Vos, Hoffmann, and Theodorou]{mckinlay2026understanding}
Jack McKinlay, Marina De~Vos, Janina Hoffmann, and Andreas Theodorou.
\newblock Understanding the process of human-{AI} value alignment.
\newblock \emph{Journal of Artificial Intelligence Research}, 85, 2026.

\bibitem[{METR}(2026)]{metr2026huggingface}
{METR}.
\newblock Brief independent investigation of agents' behavior, reasoning and collaboration in the {OpenAI} / {Hugging Face} hacking incident.
\newblock \url{https://metr.org/blog/2026-08-26-openai-hugging-face-incident-investigation/}, August 2026.

\bibitem[Motwani et~al.(2024)Motwani, Baranchuk, Strohmeier, Bolina, Torr, Hammond, and de~Witt]{motwani2024secret}
Sumeet~R Motwani, Mikhail Baranchuk, Martin Strohmeier, Vijay Bolina, Philip~H Torr, Lewis Hammond, and Christian~S de~Witt.
\newblock Secret collusion among {AI} agents: Multi-agent deception via steganography.
\newblock \emph{Advances in Neural Information Processing Systems}, 37:\penalty0 73439--73486, 2024.

\bibitem[Nakamura et~al.(2026)Nakamura, Kumar, Das, Abdelnabi, Mahmud, Fioretto, Zilberstein, and Bagdasarian]{nakamura2026colosseum}
Mason Nakamura, Abhinav Kumar, Saswat Das, Sahar Abdelnabi, Saaduddin Mahmud, Ferdinando Fioretto, Shlomo Zilberstein, and Eugene Bagdasarian.
\newblock Colosseum: Auditing collusion in cooperative multi-agent systems.
\newblock \emph{arXiv preprint arXiv:2602.15198}, 2026.

\bibitem[Narayanan \& Kapoor(2025)Narayanan and Kapoor]{narayanan2025ai}
Arvind Narayanan and Sayash Kapoor.
\newblock {AI} as normal technology.
\newblock \emph{Knight First Amendment Institute}, 25, 2025.

\bibitem[Ngo et~al.(2024)Ngo, Chan, and Mindermann]{ngo2024alignment}
Richard Ngo, Lawrence Chan, and S{\"o}ren Mindermann.
\newblock The alignment problem from a deep learning perspective.
\newblock In \emph{International Conference on Learning Representations}, volume 2024, pp.\  7474--7501, 2024.

\bibitem[{OpenAI}(2026)]{openai2026misalignment}
{OpenAI}.
\newblock Our framework for reporting model misalignment.
\newblock \url{https://openai.com/index/model-misalignment-reporting-framework/}, September 2026.
\newblock Published September 16, 2026.

\bibitem[Ouyang et~al.(2022)Ouyang, Wu, Jiang, Almeida, Wainwright, Mishkin, Zhang, Agarwal, Slama, Ray, et~al.]{ouyang2022training}
Long Ouyang, Jeffrey Wu, Xu~Jiang, Diogo Almeida, Carroll Wainwright, Pamela Mishkin, Chong Zhang, Sandhini Agarwal, Katarina Slama, Alex Ray, et~al.
\newblock Training language models to follow instructions with human feedback.
\newblock \emph{Advances in neural information processing systems}, 35:\penalty0 27730--27744, 2022.

\bibitem[Perdomo et~al.(2020)Perdomo, Zrnic, Mendler-D{\"u}nner, and Hardt]{perdomo2020performative}
Juan Perdomo, Tijana Zrnic, Celestine Mendler-D{\"u}nner, and Moritz Hardt.
\newblock Performative prediction.
\newblock In \emph{International conference on machine learning}, pp.\  7599--7609. PMLR, 2020.

\bibitem[Perez-Ortiz(2026)]{perez-ortiz2026position}
Maria Perez-Ortiz.
\newblock Position: {AI} must become planet-centered, not just human-centered.
\newblock In \emph{Forty-third International Conference on Machine Learning Position Paper Track}, 2026.
\newblock URL \url{https://openreview.net/forum?id=KqQqh2EuSG}.

\bibitem[Rafailov et~al.(2023)Rafailov, Sharma, Mitchell, Manning, Ermon, and Finn]{rafailov2023direct}
Rafael Rafailov, Archit Sharma, Eric Mitchell, Christopher~D Manning, Stefano Ermon, and Chelsea Finn.
\newblock Direct preference optimization: Your language model is secretly a reward model.
\newblock \emph{Advances in neural information processing systems}, 36:\penalty0 53728--53741, 2023.

\bibitem[Rein \& Kaufman(2026)Rein and Kaufman]{rein2026redteaming}
David Rein and Adam Kaufman.
\newblock Red-teaming {OpenAI}'s internal agent monitoring systems, October 2026.
\newblock URL \url{https://metr.org/blog/2026-10-22-red-teaming-openai-agent-monitoring/}.

\bibitem[Selbst et~al.(2019)Selbst, Boyd, Friedler, Venkatasubramanian, and Vertesi]{selbst2019fairness}
Andrew~D Selbst, Danah Boyd, Sorelle~A Friedler, Suresh Venkatasubramanian, and Janet Vertesi.
\newblock Fairness and abstraction in sociotechnical systems.
\newblock In \emph{Proceedings of the conference on fairness, accountability, and transparency}, pp.\  59--68, 2019.

\bibitem[Shen et~al.(2026)Shen, Zhu, Srinivasan, Sleight, Wagner~III, Matthews, Jones, and Sohl-Dickstein]{shen2026ai}
Judy~Hanwen Shen, Daniel Zhu, Siddarth Srinivasan, Henry Sleight, Lawrence~T Wagner~III, Morgan~Jane Matthews, Erik Jones, and Jascha Sohl-Dickstein.
\newblock {AI} organizations are more effective but less aligned than individual agents.
\newblock \emph{arXiv preprint arXiv:2604.10290}, 2026.

\bibitem[Sistla \& Clarke(1985)Sistla and Clarke]{sistla1985complexity}
A~Prasad Sistla and Edmund~M Clarke.
\newblock The complexity of propositional linear temporal logics.
\newblock \emph{Journal of the ACM (JACM)}, 32\penalty0 (3):\penalty0 733--749, 1985.

\bibitem[Skalse et~al.(2022)Skalse, Howe, Krasheninnikov, and Krueger]{skalse2022defining}
Joar Skalse, Nikolaus Howe, Dmitrii Krasheninnikov, and David Krueger.
\newblock Defining and characterizing reward gaming.
\newblock \emph{Advances in neural information processing systems}, 35:\penalty0 9460--9471, 2022.

\bibitem[Stickland et~al.(2025)Stickland, Michelfeit, Mani, Griffin, Matthews, Korbak, Inglis, Makins, and Cooney]{stickland2025async}
Asa~Cooper Stickland, Jan Michelfeit, Arathi Mani, Charlie Griffin, Ollie Matthews, Tomek Korbak, Rogan Inglis, Oliver Makins, and Alan Cooney.
\newblock Async control: Stress-testing asynchronous control measures for {LLM} agents.
\newblock \emph{arXiv preprint arXiv:2512.13526}, 2025.

\bibitem[Truong et~al.(2026)Truong, Huang, Inan, Faulkner, Christoph, Zhang, Piedrahita, and Jin]{truong2026multi}
Van~Q Truong, Xuanqiang~Angelo Huang, Erivan Inan, Ryan Faulkner, Joel Christoph, Terry~Jingchen Zhang, David~Guzman Piedrahita, and Zhijing Jin.
\newblock Multi-agent {AI} systems need institutional design, not just model-level alignment.
\newblock In \emph{Trustworthy AI for Good (AI4GOOD) Workshop@ ICML 2026}, 2026.

\bibitem[West \& Aydin(2025)West and Aydin]{west2025ai}
Robert West and Roland Aydin.
\newblock The {AI} alignment paradox.
\newblock \emph{Communications of the ACM}, 68\penalty0 (3):\penalty0 24--26, 2025.

\bibitem[Wiener(1960)]{wiener1960some}
Norbert Wiener.
\newblock Some moral and technical consequences of automation: As machines learn they may develop unforeseen strategies at rates that baffle their programmers.
\newblock \emph{Science}, 131\penalty0 (3410):\penalty0 1355--1358, 1960.

\bibitem[Wolf et~al.(2023)Wolf, Wies, Avnery, Levine, and Shashua]{wolf2023fundamental}
Yotam Wolf, Noam Wies, Oshri Avnery, Yoav Levine, and Amnon Shashua.
\newblock Fundamental limitations of alignment in large language models.
\newblock \emph{arXiv preprint arXiv:2304.11082}, 2023.

\bibitem[Xiang et~al.(2025)Xiang, Zheng, Li, Hong, Li, Xie, Zhang, Xiong, Xie, Bastian, et~al.]{xiang2025guardagent}
Zhen Xiang, Linzhi Zheng, Yanjie Li, Junyuan Hong, Qinbin Li, Han Xie, Jiawei Zhang, Zidi Xiong, Chulin Xie, Nathaniel~D Bastian, et~al.
\newblock Guardagent: safeguard {LLM} agents via knowledge-enabled reasoning.
\newblock In \emph{ICML 2025 workshop on computer use agents}, 2025.

\bibitem[Yao(2025)]{yao2025alignment}
Jasper Yao.
\newblock The alignment trap: Complexity barriers.
\newblock \emph{arXiv preprint arXiv:2506.10304}, 2025.

\bibitem[Zhao et~al.(2026)Zhao, Hu, Sui, Li, Deng, Zhao, Qin, and Che]{zhao2026gains}
Weixiang Zhao, Yulin Hu, Xingyu Sui, Zhuojun Li, Yang Deng, Yanyan Zhao, Bing Qin, and Wanxiang Che.
\newblock The gains do not make up for the losses: a comprehensive evaluation for safety alignment of large language models via machine unlearning.
\newblock \emph{Frontiers of Computer Science}, 20\penalty0 (2):\penalty0 2002319, 2026.

\bibitem[Zhi-Xuan et~al.(2025)Zhi-Xuan, Carroll, Franklin, and Ashton]{zhi2025beyond}
Tan Zhi-Xuan, Micah Carroll, Matija Franklin, and Hal Ashton.
\newblock Beyond preferences in {AI} alignment: T. zhi-xuan et al.
\newblock \emph{Philosophical Studies}, 182\penalty0 (7):\penalty0 1813--1863, 2025.

\end{thebibliography}
\bibliographystyle{iclr2027_conference}


\end{document}